\documentclass[conference]{IEEEtran}
\IEEEoverridecommandlockouts
\usepackage{cite}
\usepackage{amsmath,amssymb,amsfonts}
\usepackage{graphicx}
\usepackage{textcomp}
\usepackage{xcolor}
\usepackage{multirow}
\usepackage{booktabs}
\usepackage{hyperref}
\usepackage{tikz}
\usepackage{comment}
\usepackage{algorithm}% http://ctan.org/pkg/algorithm
\usepackage[noend]{algpseudocode}% http://ctan.org/pkg/algorithmicx
\usepackage{subcaption}
\usepackage{threeparttable}
\usepackage[font=small,labelfont=bf]{caption}
\usepackage{authblk}
\usepackage{fancyhdr}

\definecolor{ao(english)}{rgb}{0.0, 0.5, 0.0}

\newcommand{\green}[1]{\textcolor{ao(english)}{#1}} 
\newcommand{\red}[1]{\textcolor{red}{#1}} 
\newcommand{\blue}[1]{\textcolor{blue}{#1}} 
\hypersetup{
    colorlinks=true,
    linkcolor=blue,
    filecolor=magenta,      
    urlcolor=cyan,
    citecolor = magenta,
    pdfpagemode=FullScreen,
}

\usepackage{color}

\makeatletter
\newcommand*\titleheader[1]{\gdef\@titleheader{#1}}
\AtBeginDocument{%
  \let\st@red@title\@title
  \def\@title{%
\vskip-0.25em\bgroup\normalfont\large\centering\@titleheader\par\egroup
    \vskip0.5em\st@red@title}
}
\makeatother

\def\BibTeX{{\rm B\kern-.05em{\sc i\kern-.025em b}\kern-.08em
    T\kern-.1667em\lower.7ex\hbox{E}\kern-.125emX}}

\title{
 \fontsize{22.5}{30.0}\selectfont{BERRY:  \underline{B}it  \underline{E}rror  \underline{R}obustness for Energy-Efficient   \underline{R}einforcement Learning-Based Autonomous S\underline{y}stems\vspace{-3pt}}
}
\titleheader{2023 60th IEEE/ACM Design Automation Conference (DAC)}

\author[$1$]{Zishen~Wan}
\author[$2$]{Nandhini~Chandramoorthy}
\author[$2$]{Karthik~Swaminathan}
\author[$2$]{Pin-Yu~Chen}
\author[$3$]{\\Vijay Janapa Reddi}
\author[$1$]{Arijit Raychowdhury\vspace{-3pt}} 

\affil[$ $]{\normalsize \textit{$^{1}$Georgia Institute of Technology\hspace{0.05in} $^{2}$IBM Research\hspace{0.05in} $^{3}$Harvard University}\vspace{-7pt}}
\begin{document}
\maketitle

\begin{abstract}
%[Draft] Autonomous systems need to operate energy-efficiently due to their power and resource constraints. The low-voltage operation allows for further energy consumption reduction, however, causes memory bit-failures. This paper proposes a random bit-error-training technique, \textsc{BERRY}, that improves the bit-error robustness in reinforcement learning-enabled autonomous systems. \textsc{BERRY} leads to high energy savings in both compute-level operation and system-level quality-of-flight. \textsc{BERRY} generalizes across operating voltages and accelerators, as demonstrated on bit errors from profiled SRAM arrays. Experiments on various realistic virtual environments and real drone platform show that \textsc{BERRY} significantly improves bit error robustness, saves up to xx\% flight energy usage, and increases up to xx\% the number of missions.
%Various envs, both offline and online, comprehensive evaluation on robustness, processing efficiency and quality-of-flight metrics

% Aided by a marked increase in their on board processing power, state-of-the-art autonomous systems, such as  Unmanned Aerial Vehicles (UAVs) are expected to run highly complex Reinforcement-Learning (RL) models for obstacle detection and navigation, within tight power and weight constraints. 
Autonomous systems, such as Unmanned Aerial Vehicles (UAVs), are expected to run complex reinforcement learning (RL) models to execute fully autonomous position-navigation-time tasks within stringent onboard weight and power constraints. 
We observe that reducing onboard operating voltage can benefit the energy efficiency of both the computation and flight mission, however, it can also result in on-chip bit failures that are detrimental to mission safety and performance.
% While reducing the onboard supply voltage can not only improve the energy efficiency of computation, but can also impact the overall flight energy, it can also results in bit-level failures, particularly in the on-chip memories.
% not only improve the energy efficiency of computation, but impacts the overall flight energy, however, it can also results in bit-level failures, particularly in the on-chip memories. 
To this end, we propose \textsc{BERRY}, \emph{a robust learning framework to improve bit error robustness and energy efficiency for RL-enabled autonomous systems}. \textsc{BERRY} supports robust learning, both offline and on-board the UAV, and for the first time, demonstrates the practicality of robust low-voltage operation on UAVs that leads to high energy savings in both \emph{compute-level operation} and \emph{system-level quality-of-flight}. We perform extensive experiments on 72 autonomous navigation scenarios and demonstrate that \textsc{BERRY} generalizes well across environments, UAVs, autonomy policies, operating voltages and fault patterns, and consistently improves robustness, efficiency and mission performance, achieving up to 15.62\% reduction in flight energy, 18.51\% increase in the number of successful missions, and 3.43$\times$ processing energy reduction.

% \emph{a robust learning framework to improve the bit-error robustness and mission efficiency in RL-enabled autonomous systems} through a novel bit error-aware training technique, which in turn, leads to high energy savings in both \emph{compute-level operation} and \emph{system-level quality-of-flight}. \textsc{BERRY} can be applied to a wide range of operating voltages, types of drones as well as for learning both on-board the UAV and offline. Experiments on various UAV platforms, environments and autonomy policies, coupled with bit-error measurements from real SRAM chips show that \textsc{BERRY} significantly improves bit error robustness, saves 15.62\% flight energy usage, and increases 18.51\% the number of missions, with 3.43$\times$ operating energy reductions.

\begin{comment}
    berry - robust learning framework improve robustand efficiency, both compute and system
    support both offlie and ondevice robust learing
    72 scenarios deployments - prove generalization and effectivness
    highlight result
\end{comment}

\end{abstract}

% \vspace{-0.02in}
\section{Introduction}
\label{sec:intro}
% \vspace{-0.02in}

Autonomous systems are becoming prevalent for Position-Navigation-Timing (PNT) applications.
%, leading to a multitude of innovations for increased computational capability, efficiency, and reliability. 
To achieve fully autonomous PNT, state-of-the-art unmanned aerial vehicles (UAVs) are expected to run complex reinforcement learning (RL) models on-board with little-to-no offloading computation support~\cite{anwar2020autonomous,krishnan2021air,duisterhof2021tiny}.  
However, these safety-critical UAVs usually have Size, Weight, and Power (SWaP) constraints, hence it becomes imperative to deliver energy-efficient computation without compromising on robustness and safety~\cite{wan2022analyzing,wan2023vpp}. 
To satisfy these constraints, processors on-board the UAV are usually designed with techniques proposed for energy-efficient AI accelerators, such as quantization~\cite{quarl}, specialized compute units, and optimized dataflows~\cite{suleiman2019navion,wan2022circuit}. 

% Intelligent autonomous systems for autonomous navigation and Position-Navigation-Timing (PNT) are emerging rapidly leading to a multitude of innovations for increased computational capability, efficiency and reliability. State-of-the-art autonomous unmanned aerial vehicles (UAVs) navigation systems execute highly complex tasks, such as reinforcement learning based navigation, on-board with little-to-no requirement for offloading computation to a ground server.  Safety-critical autonomous systems are also highly constrained by limitations on their Size, Weight and Power (SWaP). Therefore it becomes imperative to perform computations on camera inputs to deliver real-time throughputs in an energy-efficient manner, without compromising on robustness or safety. In order to satisfy these multiple constraints, the on-board processors for UAVs can be designed with numerous techniques proposed for energy-efficient AI accelerators such as quantization, optimized dataflows and specialized/highly parallel compute units. 

In contrast to other embedded and mobile applications, the processing power is only a small fraction of the total UAV power, with the majority being used for flight motion. However, a small reduction in processing power would enable it to be re-targeted towards increasing the flight speed, thus resulting in significant energy savings on account of reduced flight time~\cite{boroujerdian2018mavbench,krishnan2022automatic}. Given the quadratic relation between energy and operating voltage, lowering the supply voltage of the onboard processor is a powerful means of energy-efficient computing within prescribed SWaP budgets. % Scaling the voltage can thus enable the entire UAV to operate faster and in a more energy-efficient manner.

\begin{figure}[t!]
\centering\includegraphics[width=\columnwidth]{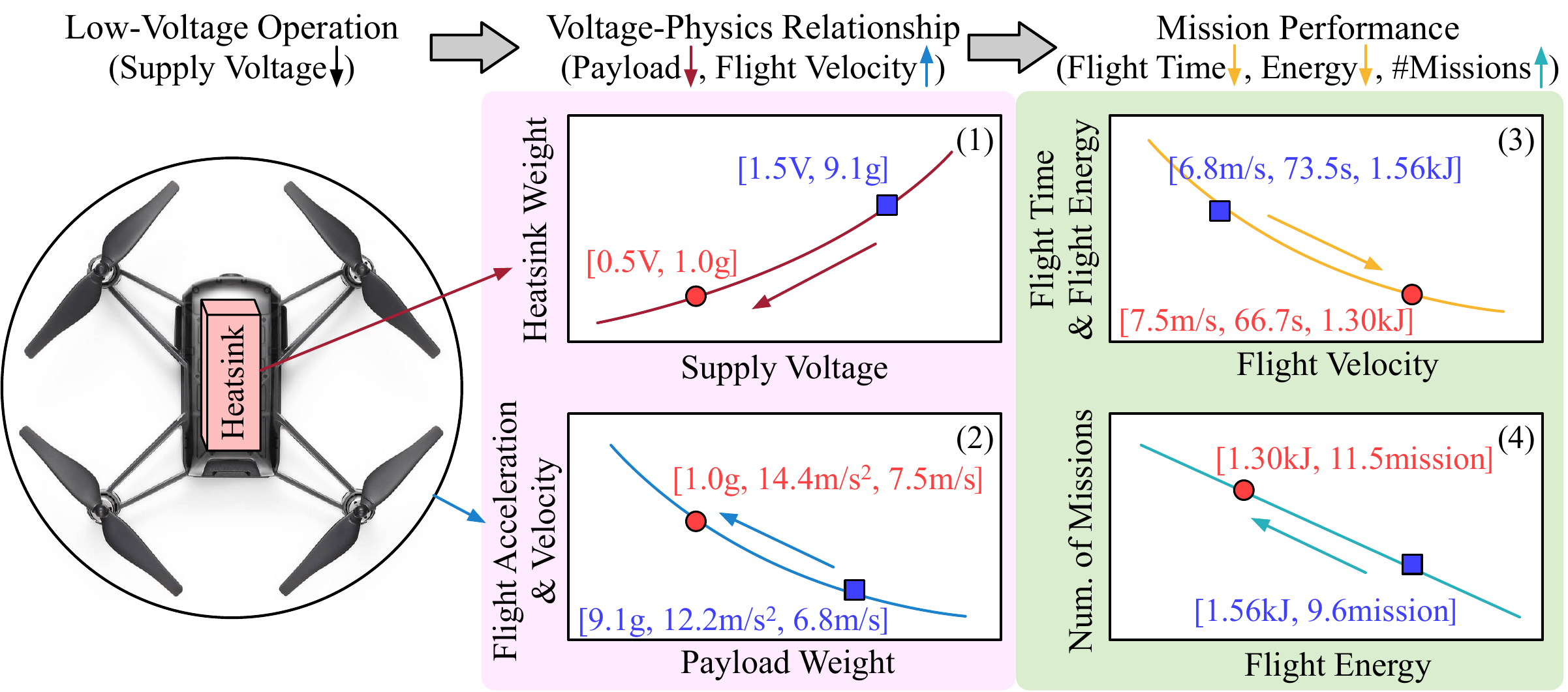}
        \caption{Relation between supply voltage, payload weight, velocity, flight time, flight energy and the number of missions observed in DJI Tello UAV. \textbf{(1)} Lowering the supply voltage of the onboard processing unit helps reduce peak temperature, and correspondingly, the heatsink weight. \textbf{(2)} A reduced payload can significantly improve acceleration and velocity, thus \textbf{(3)} reducing flight time and energy, \textbf{(4)} making the UAV complete more missions on a single battery charge.}
        \label{fig:leading}
        \vspace{-10pt}
\end{figure}

Fig.~\ref{fig:leading} shows the relationship between factors affecting UAV navigation. Scaling the processor supply voltage reduces its peak temperature, which would then reduce the required size and weight of the heatsink. This reduction in the payload translates to a further reduction in overall flight time and energy. 
A further advantage is that voltage scaling methods are complementary to and can be applied in conjunction with other energy-saving techniques described above. 

However, scaling down the voltage towards near-threshold ranges can have adverse implications on the overall reliability of the UAV. Operating below rated voltage ranges can result in memory bit errors~\cite{minervaISCA2016,amdDAC2017,danteHPCA2019, edenMICRO2019,salami2020experimental,larimi2021understanding} and logic timing errors~\cite{thundervoltDAC2018}. While~\cite{minervaISCA2016,danteHPCA2019} incur overheads of error detection and mitigation through circuit-microarchitecture methods, other works present methods to generate DNN inference models that are robust to bit errors using error-aware training~\cite{maticDATE2018,eatMLSYS2021,stutz2022random}. Although modern-day processors are equipped with RAS-enhancing features such as parity, ECC and redundancy, they are primarily targeted toward mitigating transient errors and are ineffective against low-voltage induced errors. 
%Karthik - to add why SER mitigation techniques are not enough

% In this work, we propose BERRY, a robust reinforcement learning framework for autonomous systems using the principles of error-aware training. Unlike prior work, we are the \emph{first to focus on robust learning}. In addition, we comprehensively characterize and evaluate autonomous navigation on two different drones equipped with our robust learning framework with an emphasis on quality of flight. 
In this paper, we propose \textsc{BERRY}, a robust reinforcement learning framework for autonomous systems. \textsc{BERRY} applies error-aware training to optimize system \emph{robustness}, thus boosting the processing \emph{efficiency} and improving mission-level \emph{performance} under \emph{low operating voltage}. \textsc{BERRY} supports both offline and on-device learning paradigms to enable robust and efficient low-voltage operation. \textsc{BERRY} generalizes across devices and voltages, and improves robustness and mission performance across UAVs, environments, models, and tasks.
% In this paper, we mainly use drones as platforms but the techniques can also be applied to other RL-based autonomous systems. The contributions of \textsc{BERRY} are as follows:

This paper, therefore, makes the following contributions:
\begin{itemize}
    \item We systematically analyze the impact of lowering processor voltage on autonomous systems, and determine the relation between voltage, robustness, and mission performance.
    \item We propose \textsc{BERRY}, a robust learning framework for RL-based autonomous systems, with both offline and on-device learning support. This is the \emph{first work} to focus on \emph{robust learning} for low-voltage operation on UAVs, achieving high \emph{task robustness}, \emph{processing efficiency}, and \emph{mission performance}.
    \item We evaluate \textsc{BERRY} on 72 UAV deployment scenarios and show that \textsc{BERRY} generalizes across UAVs, environments, voltages, and bit error patterns. \textsc{BERRY} achieves up to 15.62\% energy savings, 18.51\% increase in successful missions with 3.43$\times$ processing energy reduction on navigation.
\end{itemize}
\section{Background and Related Work}
\label{sec:model}

\label{subsec:fault_model}
% \textcolor{blue}{Nandhini - could you pls help this subsection and Fig.~\ref{fig:vol_ber_energy}?}
% \begin{itemize}
%     \item Fig.~\ref{fig:vol_ber_energy} center: voltage - bit error rate - SRAM energy.
%     \item Fig.~\ref{fig:vol_ber_energy} left and right: different chip bit-flip pattern.
%     \item Inclusive random bit error model.
% \end{itemize}

\begin{comment}
% \begin{figure}[t!]
%     \centering\includegraphics[width=\columnwidth]{figs/vol_ber_energy.png}
%         \caption{[Placeholder] Voltage - Bit error rate - Energy - Max frequency relation. (Need to add max frequency info)}
%         \label{fig:vol_ber_energy}
%         \vspace{-10pt}
% \end{figure}

% \begin{figure*}[t!]
%     \centering\includegraphics[width=2\columnwidth]{figs/chip_fault_pattern.png}
%         \caption{[Placeholder] Exemplary SRAM Bit Error Patterns.}
%         \label{fig:fault_pattern}
%         \vspace{-10pt}
% \end{figure*}

\end{comment}

\subsection{Reinforcement Learning (RL)-Based Autonomous Systems}
\label{subsec:autonav}
In RL-based autonomous systems, the agent learns a policy by interacting with the environment to achieve defined goals. The learning procedure is modeled by a Markov Decision Process (MDP) as $\mathcal{M} = (\mathcal{S},\mathcal{A},\mathcal{P},\mathcal{R},\gamma)$, where 
$\mathcal{S}$ is the state space, $\mathcal{A}$ is the action space, $\mathcal{P}: \mathcal{S}\times \mathcal{A}\rightarrow \mathcal{S}$ is the MDP transition probabilities, $\mathcal{R}:\mathcal{S}\times\mathcal{A}\rightarrow \mathbb{R}$ is the reward.
At each interaction $i$, the agent observes the tuple $\mathcal{D}_i = (s_i, a_i, s_{i+1}, r_i)$, where $s_i, s_{i+1}\in\mathcal{S}$ is the current and next state, $a_i \in \mathcal{A}$ is the action taken at step $i$, and $r_i = \mathcal{R}(s_i, a_i)$ is the obtained reward.  

We aim to learn an optimal policy $\pi^*$ given the observed $\mathcal{D}_i$ that can maximize the reward, i.e., $\pi^{*}(s) = \mathop{\arg\!\max}_{a} Q^{*}(s, a)$, with the function $Q^{*}: \mathcal{S}\times\mathcal{A}\rightarrow \mathbb{R}$. We use Q-learning where the Bellman backup operator is used to update the $Q$ function:
\begin{align}
    Q^{\pi}\left(s_i, a_i\right)\leftarrow \left[\mathcal{R}(s_i, a_i)+\gamma \max _{a^{\prime}} Q^{\pi}\left(s_{i+1}, a^{\prime}\right)\right]
    \label{eq:bellman}
\end{align}

The above policy converges to an optimal $\pi^*$ (deterministic) under the Bellman backup operator. We use Deep Q-Network (DQN) approximation $f_{\theta}$: $\mathcal{S}\rightarrow\mathcal{A}$ to estimate the $Q$ function parameterized by weights $\theta$. 
This neural network learns an updated mapping from $s \rightarrow Q(s,\cdot)$ using back-propagation by minimizing the loss between the predicted and Bellman updated target $Q$-values (Eq. \ref{eq:bellman}). Prior works have proved that DQN performs well in UAV autonomous systems~\cite{anwar2020autonomous,krishnan2021air}.

\subsection{Low-Voltage Induced Bit Errors}
\label{subsec:low_vol_fault}
\begin{figure}
\includegraphics[width=\columnwidth]{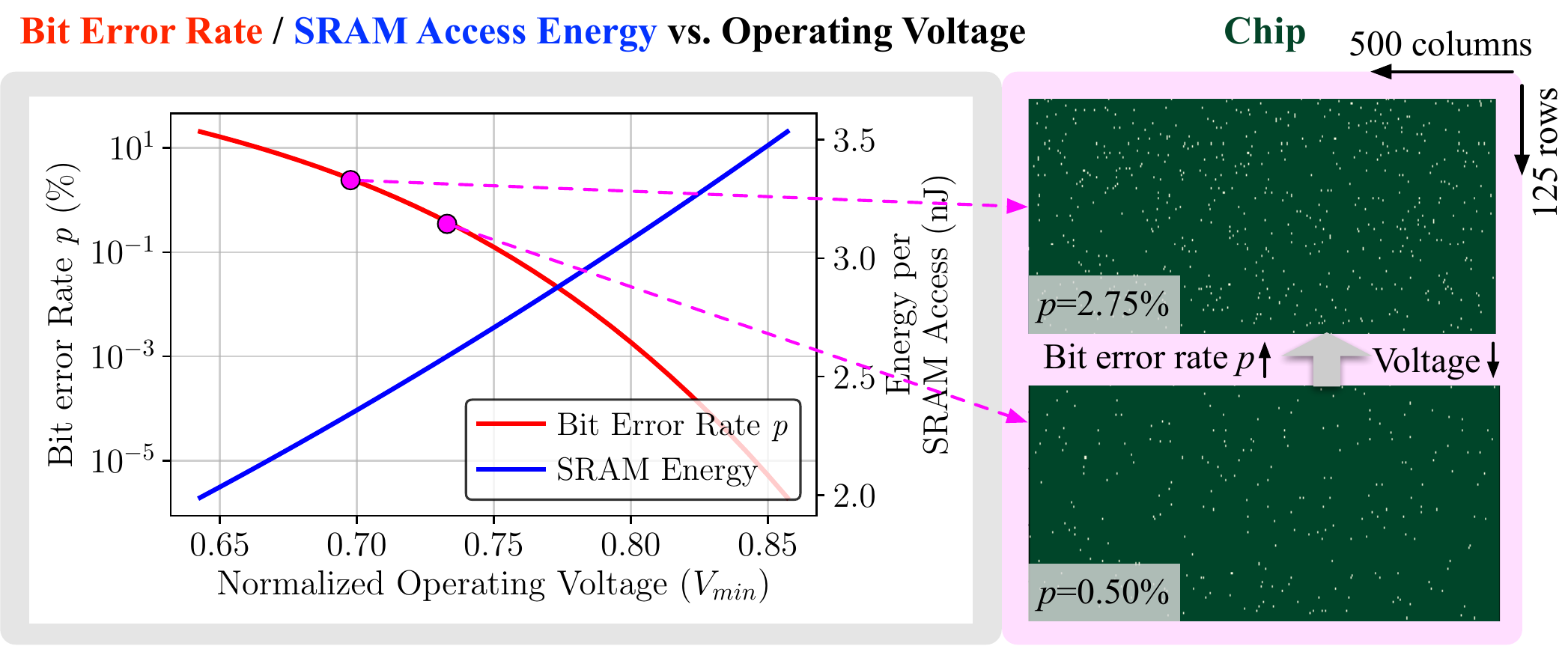}
        \caption{\textbf{Low-Voltage Operation, Energy and Bit Errors.} SRAM bit error rate increases and access energy drops with decreasing supply voltage for a 14nm FinFET SRAM chip fabricated in~\cite{danteHPCA2019}. The voltage (x-axis) is normalized to $V_{min}$, the lowest measured voltage at which there are no bit errors. Reproduced on the right is a random spatial error pattern in a cross-section of the memory array from~\cite{danteHPCA2019,eatMLSYS2021}.}
        \label{fig:vol_ber_energy}
        \vspace{-10pt}
\end{figure}

Lowering operating voltages towards near-threshold ranges exacerbates bit cell variations.
This manifests as an exponential increase in bit errors, affecting accelerator memories in which weights are stored and updated. Fig.~\ref{fig:vol_ber_energy} shows this relationship for an exemplary 14nm FinFET SRAM chip fabricated in~\cite{danteHPCA2019} and a sample spatial distribution of bit errors in a segment of the tested memory arrays. At a given voltage, these bit flips are persistent across multiple reads and write to the same location. The locations are random and independent of each other across different chips and arrays~\cite{maticDATE2018,amdDAC2017,eatMLSYS2021}. 
It needs to be noted that these are not transient errors, so computational redundancy in space and time cannot mitigate them. Similarly, standard ECC may not correct all observed errors since there could be multiple faulty bits per memory word. %Sec.~\ref{sec:characterization} explores the impact of such bit errors in navigation and learning on autonomous systems in detail.

\subsection{Bit Error-Aware Training}
% To improve bit error robustness, researchers have developed various error mitigation techniques. Error
Rather than incurring the cost of hardware-based error mitigation~\cite{minervaISCA2016,minervaISSCC2017,danteHPCA2019}, some works~\cite{maticDATE2018,eatMLSYS2021,edenMICRO2019} advocate improving robustness by generating a model resilient to bit errors. 
Profiled bit errors are injected during offline training on error-free hardware, resulting in a robust model during inference on low-voltage devices. \textsc{BERRY} proposes the generation of robust models but tackles entirely different challenges: (i) \textsc{BERRY} targets learning (both offline and on-device) as opposed to inference, meaning that \emph{learning can occur on low-voltage devices} with bit errors affecting parameters. (ii) Prior works are restricted to supervised object classification while \textsc{BERRY} focuses on \emph{reinforcement learning}, leading to a new error-aware training framework. (iii) \textsc{BERRY} tackles the complex relationship between low-voltage operation and cyber-physical autonomous systems aiming to \emph{improve both compute-level and system-level efficiency while ensuring robustness}.

% As described in~\cite{amdDAC2017, danteHPCA2019},  lowering operating voltages towards near-threshold operation further exacerbates bit cell variations in 14nm FinFET SRAMs. This manifests as an exponential increase in bit errors with lowered supply voltage, affecting accelerator memories in which weights are stored and updated. Figure~\ref{fig:vol_ber_energy} shows this relationship for two chips fabricated in~\cite{danteHPCA2019} and  a sample spatial distribution of bit errors in a segment of the tested memory arrays. Prior work~\cite{maticDATE2018,amdDAC2017,eatMLSYS2021} assumes that at a given voltage, these bit error locations are random and independent of each other in a memory array and across different chips and arrays. Section~\ref{sec:characterization} explores the impacts of such bit errors in navigation and learning on autonomous vehicles in detail.

% \begin{figure}
% \includegraphics[width=0.9\columnwidth]{figs/background-ber.png}
%         \caption{SRAM bit error rate increase with decreasing supply voltage for 2 different chips fabricated in~\cite{danteHPCA2019} in 14nm technology. The voltage (x-axis) is normalized to $V_{min}$, the lowest measured voltage at which there are no bit errors. Reproduced on the right is a random spatial error pattern in a cross-section of the memory array from~\cite{danteHPCA2019,eatMLSYS2021}}
%         \label{fig:vol_ber_energy}
%         \vspace{2pt}
% \end{figure}

\subsection{Resilience Characterization in UAVs}
Several prior works have attempted to characterize the resilience of applications running on-board UAVs.
% ~\cite{venkatagiri2018swapprox} analyzes software approximation impacts on the resilience of a video-summarization algorithm running on a UAV processor.
\cite{wan2021analyzing,wan2022frl,hsiao2023mavfi} analyze the transient error impact on the resilience of learning-based and control-based UAV navigation systems. \cite{chan2020rlreliability,zhang2021robust} characterize the reliability and robustness of RL algorithms in terms of their inter-run variability, while~\cite{yang2022resilientQlearning} proposes techniques to train robust RL models in the presence of adversarial perturbations and interferences. In contrast, \textsc{BERRY} tackles the problem of bit errors due to low voltages while running RL models on UAVs, and supports both offline and on-device robust learning to mitigate fault effects. 
% Unlike our proposed technique, none of these works tackle the problem of bit errors due to low voltages while running RL algorithms on UAVs or propose error-aware training techniques to mitigate their effects.

% \begin{itemize}
%     % \item General RL DQN(Fig, formulate equations, etc). Specific drone autonomous navigation problem. Prior works~\cite{anwar2020autonomous,krishnan2021air} have shown that DQN works well on high-level autonomous navigation tasks for aerial robots.
%     \item Policy \#parameter: 1.27MB. We assume all NN weights are stored on-chip SRAM.
%     \item We use continuous learning and train/test environments to make trained policy be realistic. The implementation details are beyond the scope of this paper and interested readers can refer to~\cite{anwar2020autonomous,krishnan2021air}.
% \end{itemize}

\begin{figure}[t!]
% \vspace{-5pt}
\centering\includegraphics[width=.9\columnwidth]{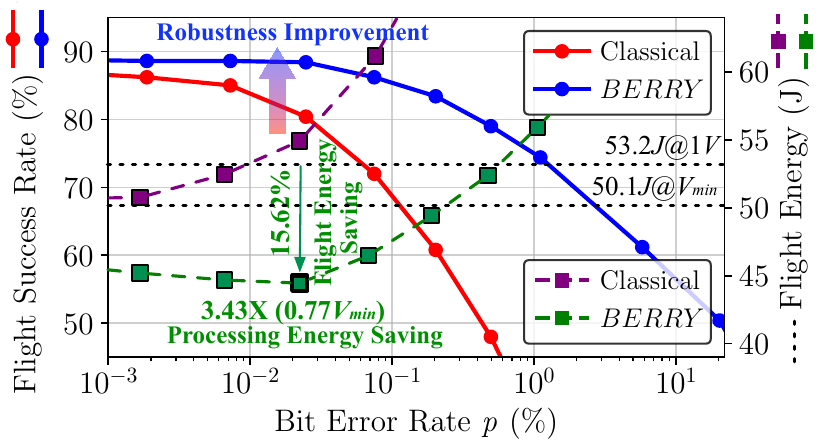}
% \vspace{-5pt}
        \caption{\textbf{Robustness to Bit Errors and Flight Energy Savings.} Low-voltage-induced bit errors degrade flight success rate, while \textsc{BERRY} improves robustness significantly. Robustness to higher bit error rates allows more energy-efficient operation (Fig.~\ref{fig:vol_ber_energy}). Compute correlates with physics in UAV (Fig.~\ref{fig:leading}), and the robust operation brings 15.62\% flight energy savings and 18.51\% more missions for navigation tasks.}
        \label{fig:succ_rate}
        \vspace{-5pt}
\end{figure}

\section{Low Voltage Fault Impact Characterization}
\label{sec:characterization}

This section explores how low-voltage operation and its induced bit errors impact the robustness, efficiency, and mission performance of autonomous navigation systems.

We evaluate the impact of low-voltage operation on an example autonomous system where the UAV aims to navigate from the start position to the goal position in the shortest time without colliding into obstacles (Sec.~\ref{subsec:setup}). Task success rate, processing energy, and quality-of-flight (flight time, flight energy, endurance) are evaluated during the mission. 

\textbf{Task success rate:} As in Fig.~\ref{fig:succ_rate}, we observe that the low-voltage induced bit errors greatly degrade autonomous system robustness. The increasing number of bit errors gradually pollutes error-free learned policy and results in the UAV taking the wrong actions to collide with obstacles. The task success rate drops to $<$85\% when bit error rate $p>0.01\%$ ($\sim$0.8$V_{min}$).
  
\textbf{Flight time:} The corrupted flight actions due to bit errors can directly lead to path detours, resulting in longer trajectory distance and extended flight time for a given task. 
  
\textbf{Flight energy:} The increased flight time consequently increases total flight energy despite processing energy savings from low-voltage operation (Fig.~\ref{fig:succ_rate}). This is because $\sim$95\% flight energy is consumed by rotors that closely correlate with the flight time.    
  
\textbf{Endurance:} The increased single-mission flight energy and duration further reduces the total number of missions that the UAV can successfully complete on a single charge before its battery depletes.
% Under a certain battery capacity, the increased flight energy means the drone finishes less number of mission
% The degrading in task success rate consequently result in quality-of-flight
% fault impact: (1) succrate drop and failure cases, (2) trajectory detour result in longer flight time, thus higher flight energy in spite of operation energy savings, (3) less number of missions.

To achieve the robustness of an autonomous system under low voltages and improve its processing efficiency and quality-of-flight, we propose \textsc{BERRY}, using principles of error-aware training.
Fig.~\ref{fig:succ_rate} highlights the key results of \textsc{BERRY} on the same UAV navigation system: with a drop in success rate of $<$1\%, 3.43$\times$ processing energy savings,  15.62\% single-mission flight energy savings, and 18.51\% more number of missions are achieved (compared with normal 1$V$ operation), with simply lowering supply voltage to 0.77$V_{min}$. %\textsc{BERRY} supports both offline and on-device robust learning. The quality-of-flight improvements and bit error robustness learned through \textsc{BERRY} generalize across chips, voltages, UAVs, autonomy policies, and environments.

\begin{figure}[t!]
% \vspace{-8pt}
\centering\includegraphics[width=\columnwidth]{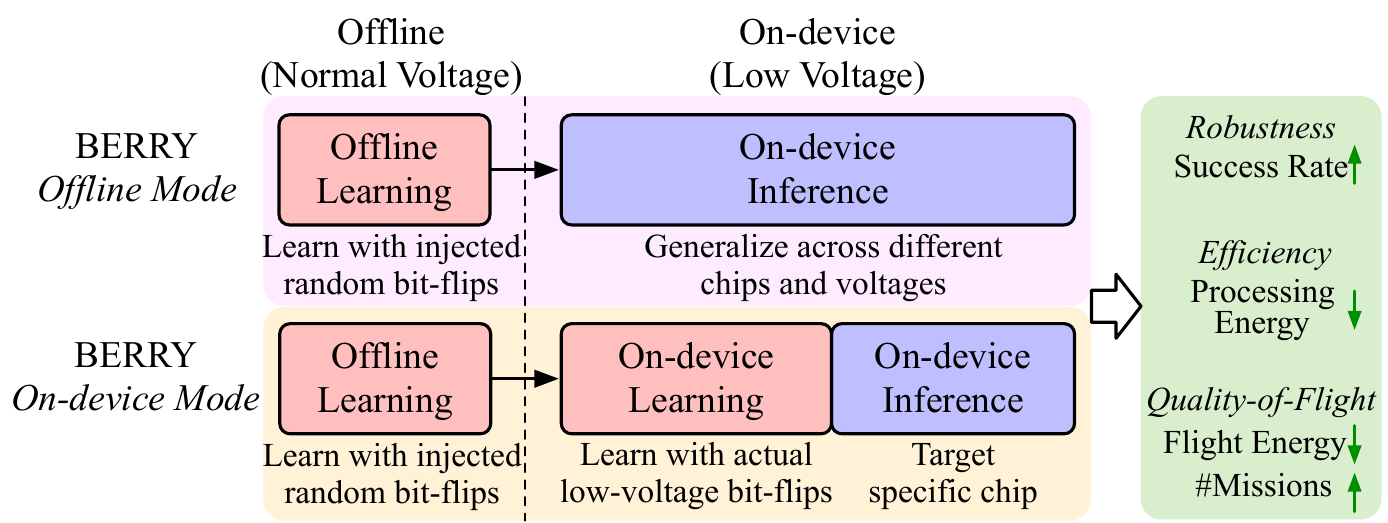}
        \caption{\textbf{\textsc{BERRY} with Offline and On-Device Robust Learning Modes}. \textsc{BERRY} offline learning with random bit-flips yields generalizable robustness and quality-of-flight improvements across voltages and chips. \textsc{BERRY} on-device learning with specific error patterns enables even lower voltage with higher robustness and efficiency.}
        \label{fig:berry_diagram}
        \vspace{2pt}
\end{figure}

\section{\textsc{BERRY} Robust Learning Framework}
\label{sec:training}
\vspace{-11pt}
This section presents \textsc{BERRY} robust reinforcement learning framework, supporting both offline and on-device learning paradigms for robustness and efficiency improvements (Fig.~\ref{fig:berry_diagram}). 
For systems without on-device learning capabilities, \textsc{BERRY} conducts learning offline with injected bit-flips at nominal voltage, and then deploys the robust policy on low-voltage SWaP-constrained UAV systems. For systems with on-device learning capabilities~\cite{anwar2020autonomous}, \textsc{BERRY} enables the UAV to learn a robust policy directly on the low-voltage device incurring errors, and then perform the navigation task on the same hardware. Combining learning and navigation on the same device ensures improved success rates and higher flight energy savings tailored to the specific chip. In both cases, we perform a round of offline learning with injected bit errors, which provides generalized robustness across devices. 

Algo.~\ref{algo:rlbet} describes the main component of \textsc{BERRY}, the robust learning methodology for RL-based autonomous systems. The objective is to learn a robust state action-value function of policy $Q(\theta)$. In standard Deep-Q-Learning, an evaluation network predicts the Q-value at each step, and the target network computes the Bellman temporal difference (Eq.~\ref{eq:bellman}), with both having the same architectures. Lines 2-3 initialize both these networks. For each step $t$ of an episode, a minibatch of state-action-reward inputs is chosen from the experience replay buffer (line~10), and the evaluation and target networks compute the predicted Q-value and the temporal difference target $y_j$ (line~12), respectively. The loss function is the expected squared difference between the two, and gradient $\Delta^{(t)}$ with respect to parameters $\theta$ is computed (line~13). While the Q function network parameters $\theta$ are updated every step, the target network parameters $\theta^{-}$ are updated every $C$ steps to keep the training stable by copying over $\theta$ to $\theta^{-}$ (line~21).

\textbf{Fault injection:} In \textsc{BERRY}, both the Q-function and target networks are injected with bit errors. A random distribution of bit error locations is generated using Voltage-BER curves from existing memory characterization (Sec.~\ref{subsec:low_vol_fault}). Bit flips from 1-to-0 and 0-to-1 are both injected at a given voltage/bit error rate, affecting both weights and activations that are stored in the on-chip SRAM. Line 15 refers to error injection following per-layer 8-bit quantization with rounding in $\theta$ and $\theta^{-}$, to obtain perturbed $\tilde{\theta}$ and $\tilde{\theta}^{-}$, respectively. 

\setlength{\textfloatsep}{3pt}% Remove \textfloatsep
\begin{algorithm}[t]
% \footnotesize
\small
  \caption{\textsc{BERRY} Robust Error-Aware Training Framework for Reinforcement Learning-Based Autonomous Systems}\label{algo:rlbet}
  \begin{algorithmic}[1]
    \Procedure{\textsc{BERRY}}{$p$}
    \State Initialize action-value function $Q$ with weight $\theta$
    \State Initialize target action-value function $\hat{Q}$ with weight $\theta^-=\theta$
\For{\textit{episode} $e = 1$ \textit{to} $E$}
      \For{\textit{time step} $t = 1$ \textit{to} $T$}
    \State  Given state $s_t$, take action $a_t$ based on $Q$ ($\epsilon$-greedy)
    \State  Obtain reward $r_t$ and reach new state $s_{t+1}$
      
        \State Store transition $(s_t, a_t, r_t, s_{t+1})$ in $D$
      % \State \{Experience replay:\}
      \State \textit{\blue{// Experience replay}}
      \State Sample a mini-batch $\{(s_j, a_j, r_j, s_{j+1})\}^{B}_{b=1}$ from $D$
      
        \State \textit{\blue{// Clean training pass}} \label{algo:clean}
        \State Set $y_j = r_j + \gamma \max_{a^{\prime}} Q(s_{j+1}, a^{\prime}; \theta^{-(t)})$
        % \State $\delta(\theta^{(t)}) = (Q(s_j, a_j; \theta^{(t)}) - y_j)^2$ 
        \State $\Delta^{(t)} = \nabla_{\theta} \sum^{B}_{b=1} (Q(s_j, a_j; \theta^{(t)}) - y_j)^2$
        
        \State \textit{\blue{// Perturbed training pass, inject bit errors at rate $p$}}
        \State $\tilde{\theta}^{(t)} = BErr_{p}(\theta^{(t)})$
        \State Set $\tilde{y_j} = (r_j + \gamma \max_{a^{\prime}} Q(s_{j+1}, a^{\prime}; \tilde{\theta}^{-(t)}))$
        % \State $\delta(\tilde{\theta}^{(t)}) = (Q(s_j, a_j;\tilde{\theta}^{(t)}) - \tilde{y_j})^2$
        \State $\tilde{\Delta}^{(t)} = \nabla_{\theta} \sum^{B}_{b=1} (Q(s_j, a_j;\tilde{\theta}^{(t)}) - \tilde{y_j})^2$ \label{algo:perturb}

        \State \textit{\blue{// Average gradients and update w.r.t $\theta$}}
        \State $\theta^{(t+1)} = \theta^{(t)} - \alpha(\Delta^{(t)} + \tilde{\Delta}^{(t)})$
        
        \State \textit{\blue{// Periodic update of target network}}
        \State Every $C$ steps reset $\hat{Q} = Q$, i.e., set $\theta^{-} = \theta$
    \EndFor
  \EndFor \EndProcedure\\
    \textbf{Output:} Bit-error robust action-value function $Q(\theta)$   
  \end{algorithmic}
\end{algorithm}

\textbf{Gradient update:}
$Q$ and $y_j$ are computed with perturbed parameters $\tilde{\theta}$ and $\tilde{\theta}^{-}$ (line~16). The new loss, termed perturbed loss, is computed along with the gradient $\tilde{\Delta}^{(t)}$ (line 17). Robust learning in \textsc{BERRY} is designed to work in both error-free and faulty hardware with voltage scaling without having to retrain the model. Therefore, the parameter update (line 19) uses an average of the perturbed and unperturbed gradients in stochastic gradient descent.  
\section{\textsc{BERRY} Evaluation}
\label{sec:evaluation}
This section evaluates \textsc{BERRY} on UAV autonomous navigation systems with 72 different scenarios. We demonstrate that \textsc{BERRY} consistently improves \textit{task robustness}, \textit{processing and mission-level efficiency} for both offline and on-device learning, and \textit{generalize} well across various environments, UAVs, models, and error patterns measured from real chips.

\subsection{Experimental Setup}
\label{subsec:setup}
\textbf{Simulation Platform.} We use open-source UAV simulation infrastructures~\cite{anwar2020autonomous,krishnan2021air} to evaluate \textsc{BERRY} framework. The infrastructure is powered by Unreal Engine to simulate the environments, AirSim to capture UAV’s dynamics and kinematics, and RL algorithms to generate flight commands in real time. 

\textbf{Task and Policy Model.} We adopt the autonomous navigation task (e.g., package delivery), where the UAV is initialized at a start location and navigates across the environment to reach the destination without colliding with obstacles. We use a perception-based probabilistic action space $\mathcal{A}$ with 25 actions, and C3F2 neural network policy~\cite{wan2021analyzing} with 1.1MB parameters. We assume the underlying systolic array-based architecture with on-chip SRAM, and integrate the SCALE-Sim~\cite{samajdar2020systematic} and Accelergy~\cite{wu2019accelergy} simulators with our energy plugin (Fig.~\ref{fig:vol_ber_energy}) to evaluate processing performance and energy. Along with voltage, we also scale the frequency based on measured results on a deep-learning accelerator reported in~\cite{jia_esscirc22}.

\textbf{UAV platforms.} We use a Crazyflie nano UAV~\cite{crazyflie}, with 27$g$ takeoff weight, 15$g$ max payload, 250$mAh$ battery capacity, and 7min max flight time. In Sec.~\ref{subsec:drone_type}, we configure another micro UAV DJI Tello~\cite{tello} with 80$g$ takeoff weight, 1100$mAh$ battery and 13$min$ max flight time for evaluation.

\textbf{Evaluation Metrics.} We evaluate both compute-level efficiency (processing energy) and mission-level metrics (success rate, flight time, flight energy, number of missions). Success rate is the percentage of successful trials, flight time and flight energy are the single-mission time and energy that are required for UAV to reach its goal, and the number of missions denotes the total missions that the UAV can complete on a battery charge. For each case, we evaluate 500 different fault maps and report the average quantity for all metrics.

% \begin{itemize}
%     \item CrazyFlie drone spec table, battery capacity, mass, etc
% \end{itemize}

\subsection{Robustness and Efficiency Improvements}
\label{subsec:robust_efficiency}
% \vspace{-4pt}

\textbf{Robustness Improvement.} Tab.~\ref{tab:succrate_comp} shows the navigation task success rate with different bit error rates $p$ under different voltages (Fig.~\ref{fig:vol_ber_energy}). Both classical DQN training policy and \textsc{BERRY} can reach $>$88\% success rate in error-free missions. However, on lowering supply voltage, the classical trained policy is vulnerable to induced bit failures, while \textsc{BERRY} can still achieve $\sim$80\% success rate under $p=0.5\%$ ($\sim$0.72$V_{min}$). The success rate is comparable to other reported autonomous navigation task success rates for similar difficulty levels~\cite{krishnan2021air}.

\begin{table}[t!]
% \large
\centering
\caption{\textbf{Robustness Improvement.} Average success rates under various bit error rates $p$. \textsc{BERRY} improves autonomous navigation task robustness under bit failures compared to classical RL policy.}
\renewcommand*{\arraystretch}{1.2}
\setlength\tabcolsep{5pt}
\resizebox{\linewidth}{!}{%
\begin{tabular}{l|c|ccccc}
\hline
\multirow{2}{*}{\begin{tabular}[c]{@{}l@{}}Autonomy \\ Schemes\end{tabular}} & \multirow{2}{*}{\begin{tabular}[c]{@{}c@{}}Error-Free  \\ SuccRate (\%)\end{tabular}} & \multicolumn{5}{c}{Bit-Error SuccRate (\%)} \\ \cline{3-7} 
                                                                           &                                                                           & $p$=0.01\%  & $p$=0.05\%  & $p$=0.1\% & $p$=0.5\% & $p$=1\%  \\ \hline
Classical                                                                     & 88.4                                                                      & 84.0    & 78.2    & 69.2  & 48.6  & 33   \\
\textbf{\textsc{BERRY}}                                                                      & \textbf{88.8}                                                                      & \textbf{88.6}    & \textbf{86.6}    & \textbf{84.4}    & \textbf{79.2}  & \textbf{74.8} \\ \hline
\end{tabular}
}
\label{tab:succrate_comp}
% \vspace{-3pt}
\end{table}

\begin{table*}[t!]
% \huge
% \vspace{-5pt}
\centering
\caption{\textbf{Operating and System Efficiency Improvement.}  Operating energy, task success rate, and system-level quality-of-flight under low voltages. \textsc{BERRY} improves operating energy efficiency due to quadratic effect of lowering voltage. \textsc{BERRY} improves system mission-level efficiency with reduced flight time, flight energy and more number of missions owning to low-voltage operation and improved robustness.}
\renewcommand*{\arraystretch}{1.1}
\resizebox{\linewidth}{!}{%
\begin{tabular}{cc|c|c|c|c|cc|cc}
\hline
\multicolumn{2}{c|}{\textbf{Low-Voltage Operation}}  & \textbf{Processing}                                                                           & \textbf{Robustness}                                                   & \multicolumn{6}{c}{\textbf{Autonomous System Mission-Level Quality-of-Flight}}                                                                                                                                                                                                                  \\ \hline
\begin{tabular}[c]{@{}c@{}}Voltage \\ (V)\end{tabular} & \begin{tabular}[c]{@{}c@{}}Bit Error \\ Rate $p$ (\%)\end{tabular} & \begin{tabular}[c]{@{}c@{}}Energy \\ Savings (\%)\end{tabular} & \begin{tabular}[c]{@{}c@{}}Success \\ Rate (\%)\end{tabular} & \begin{tabular}[c]{@{}c@{}}Flight \\ Distance (m)\end{tabular} & \begin{tabular}[c]{@{}c@{}}Flight \\ Time (s)\end{tabular} & \begin{tabular}[c]{@{}c@{}}Flight Energy  \\ $E_{flight}$ (J)\end{tabular} & \begin{tabular}[c]{@{}c@{}}$E_{flight}$  \\ Savings \end{tabular} & \begin{tabular}[c]{@{}c@{}}Num. of Missions \\ $N_{mission}$ \end{tabular} & \begin{tabular}[c]{@{}c@{}}$N_{mission}$ \\ Improvements \end{tabular} \\ \hline

1       & 0    & -                    & 88.4         & 14.89      & 6.81                          & 53.19     & -        & 55.35  & -          \\ \hline
0.86$V_{min}$              & 1.96$\times 10^{-6}$  & 2.77$\times$                & 88.0         & 14.93       & 6.51                         & 47.23  & \green{-11.21\%}           & 62.05   & \green{+12.12\%}         \\
0.84$V_{min}$             & 1.38$\times10^{-5}$  & 2.87$\times$               & 89.2          & 14.86        & 6.48                       & 46.66   & \green{-12.28\%}          & 63.66   & \green{+15.03\%}         \\
0.83$V_{min}$             & 8.23$\times10^{-5}$  & 2.97$\times$                & 89.0        & 14.91         & 6.46                        & 46.41 & \green{-12.73\%}            & 63.85  & \green{+15.37\%}          \\
0.81$V_{min}$             & 4.22$\times10^{-4}$  & 3.07$\times$               & 88.8        & 14.96        & 6.45                         & 46.22  & \green{-13.11\%}           & 63.98  & \green{+15.61\%}          \\
0.80$V_{min}$             & 1.87$\times10^{-3}$  & 3.18$\times$                & 88.6          & 14.94      & 6.42                         & 45.80  & \green{-13.90\%}           & 64.42  & \green{+16.40\%}          \\
0.79$V_{min}$             & 7.25$\times10^{-3}$  & 3.30$\times$                & 88.6       & 14.94        & 6.39                          & 45.38  & \green{-14.67\%}           & 65.01  & \green{+17.46\%}         \\
\textbf{0.77$V_{min}$}             & \textbf{2.47$\times10^{-2}$}   & \textbf{3.43$\times$}               & \textbf{88.4}         & \textbf{14.91}         & \textbf{6.35}                       & \textbf{44.88}  & \green{\textbf{-15.62\%}}           & \textbf{65.59}   & \green{\textbf{+18.51\%}} \\
0.76$V_{min}$             & 7.49$\times10^{-2}$  & 3.55$\times$                & 86.2       & 15.71           & 6.67                       & 46.90  & \green{-11.82\%}           & 61.20  & \green{+10.58\%}          \\
0.74$V_{min}$             & 2.03$\times10^{-1}$  & 3.69$\times$                & 83.4        & 16.58           & 7.03                      & 49.14  & \green{-7.61\%}           & 56.52  & \green{+2.12\%}          \\
0.73$V_{min}$             & 4.98$\times10^{-1}$  & 3.84$\times$                & 79.0          & 18.03        & 7.61                       & 52.98  & \green{-0.39\%}           & 49.66  & \red{-10.27\%}          \\
0.71$V_{min}$              & 1.11                & 3.99$\times$    & 74.4            & 19.46    & 8.18                         & 56.62 & \green{-6.45\%}            & 43.75    & \red{-20.95\%}        \\
0.68$V_{min}$            & 5.80                 & 4.42$\times$    & 63.2        & 21.84        & 9.09                       & 61.96 & \red{+16.49\%}           & 33.96   & \red{-38.64\%}         \\
0.64$V_{min}$            & 20.36                 & 4.93$\times$     & 50.4         & 24.52         & 10.11                     & 67.83 &  \red{+27.53\%}          & 24.74   & \red{-55.30\%}         \\\hline
\end{tabular}
}
\label{tab:rlbet_summary_v2}
\vspace{-2pt}
\end{table*}

\begin{figure*}[t]
% \vspace{-5pt}
\centering\includegraphics[width=2.03\columnwidth]{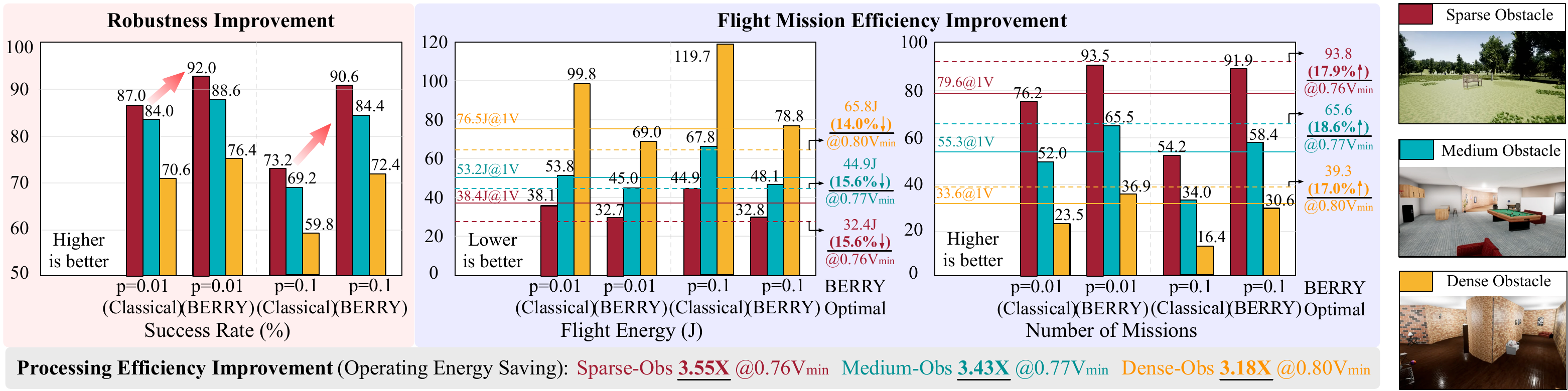}
        \caption{\textbf{Effectiveness across Different Environments}. \textsc{BERRY} is evaluated in three environments with different obstacle densities. \textsc{BERRY} consistently improves task robustness and mission efficiency (reduced single-mission flight energy and increased number of missions). \textsc{BERRY} is adaptive to various environments, and enables lower-voltage operations in sparse (0.76$V_{min}$) than complex environments (0.80$V_{min}$).}
        \label{fig:diff_env}
        \vspace{-5pt}
\end{figure*}

\begin{figure}[t!]
\centering\includegraphics[width=\columnwidth]{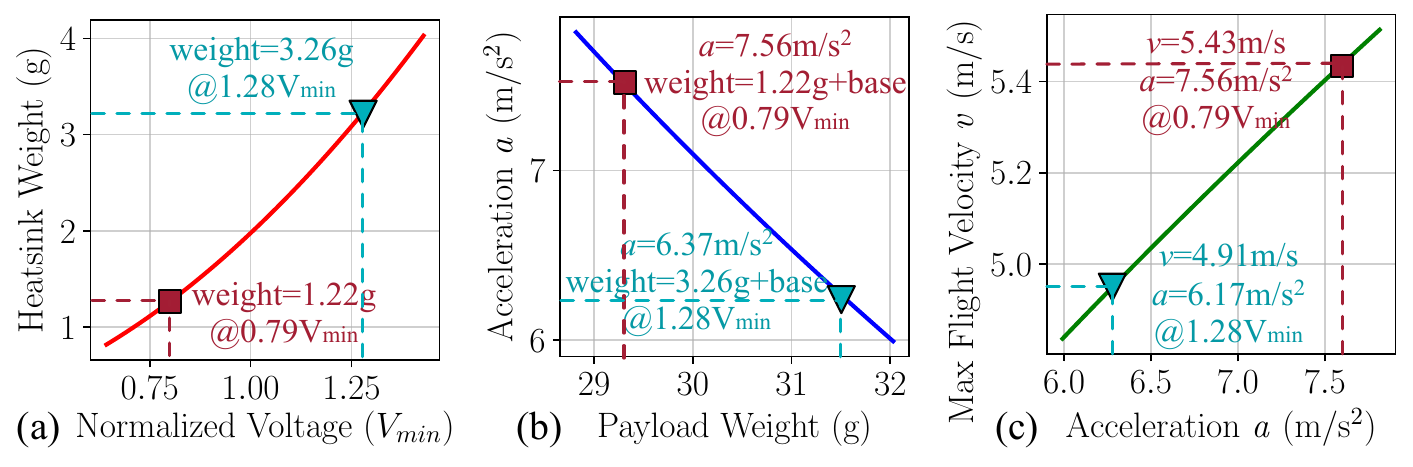}
% \vspace{-0.15in}
        \caption{\textbf{Low Operating Voltage Brings System Benefits.} \textbf{(a)} Lower operating voltage $\rightarrow$ lower energy and thermal design power (TDP) $\rightarrow$ require smaller heatsink size with reduced weight. \textbf{(b)} Lower payload weight $\rightarrow$ higher motion acceleration. \textbf{(c)} Higher acceleration $\rightarrow$ more agile when facing obstacle $\rightarrow$ higher safe flight velocity. With higher velocity, the UAV can finish a mission faster (lower flight time), thus consuming less flight energy for a single mission. This enables more missions under a single battery charge.}
        \label{fig:payload}
        % \vspace{-10pt}
\end{figure}

\textbf{Processing Efficiency Improvement and System Implication.}
Lower voltage brings quadratic energy-saving benefits. As in Tab.~\ref{tab:rlbet_summary_v2}, compared with 1$V$ normal operation~\cite{palossi201964}, lowering voltage to 0.77$V_{min}$ achieves 3.43$\times$ energy efficiency. As validated on real UAVs, the processing efficiency improvement further brings benefits to the cyber-physical UAV system~\cite{krishnan2020sky,krishnan2022roofline}. The lower energy with lower TDP (thermal design power) requires a smaller heatsink with its reduced weight~\cite{heat-sink} (Fig.~\ref{fig:payload}\blue{a}), which then yields increased motion acceleration (Fig.~\ref{fig:payload}\blue{b}). With higher acceleration, the UAV becomes more agile when facing obstacles and can have higher safe flight velocity~\cite{krishnan2020sky,krishnan2022roofline} (Fig.~\ref{fig:payload}\blue{c}). When we lower the voltage from 1.28$V_{min}$ to 0.79$V_{min}$, the operating energy reduces 2.67$\times$, with the required heatsink weight reducing from 3.26$g$ to 1.22$g$, making the UAV achieve higher acceleration (6.37$m/s^2$ to 7.56$m/s^2$) and safe flight velocity (4.91$m/s$ to 5.43$m/s$), further benefiting autonomous system mission-level efficiency.

\textbf{Mission Efficiency Improvement.} Tab.~\ref{tab:rlbet_summary_v2} shows the autonomous system mission-level performance (i.e., flight distance, time, energy, and the number of missions) with \textsc{BERRY} under low voltages. The improved robustness benefits the mission success rate maintained at $\sim$88\% and the path distance at $\sim$15$m$ under 0.77$V_{min}$. The flight distance then increases due to path detours induced by sub-optimal actions. The flight time drops to 6.35$s$ when lowering voltage to 0.77$V_{min}$ due to higher flight velocity (Fig.~\ref{fig:payload}) for the same flight distance. %The flight time then increases because of longer trajectory. 
Similarly, the flight energy reduces from 53.19$J$ to 44.88$J$ owing to the shortened flight time and the reduced power. The number of completed missions ($N$) under a battery charge ($E$) closely correlates to the success rate ($SR$) and single-mission energy ($E^{'}$) as $N=SR\times E/E^{'}$. The number of missions increases from 55 to 65 owing to the reduced flight energy and increased success rate. Overall, \textsc{BERRY} enables lower voltage operation for robust and efficient autonomous systems. At 0.77$V_{min}$ with 0.025\% $p$, \textsc{BERRY} achieves 15.62\% (10.95\%) less flight energy, 18.51\% (11.99\%) more missions with 3.43$\times$ (2.04$\times$) operating energy savings compared to 1$V$ ($V_{min}$).

% \vspace{-0.02in}
\subsection{Environment Evaluation}
\label{subsec:env}
\textbf{Effectiveness across Different Environments.} Fig.~\ref{fig:diff_env} evaluates \textsc{BERRY} on three environments with different obstacle densities, namely sparse (outdoor), medium (indoor) and dense (indoor) obstacle environments. 
% The drone conducts the same autonomous navigation task in all environment, and we record the average task success rate and mission-level quality-of-flight under various operating voltages. 
Compared with classical DQN policy, it is well observed that \textsc{BERRY} improves the success rate and mission efficiency, with 3.55$\times$, 3.43$\times$, 3.18$\times$ operating energy savings, 15.6\%, 15.6\%, 14.0\% single-mission flight energy reduction, and 17.9\%, 18.6\%, 17.0\% more number of missions for sparse, medium, dense obstacle environments, respectively (numbers underlined in Fig.~\ref{fig:diff_env}). Compared within three environments, \textsc{BERRY} enables lower operating voltage in sparse obstacle (0.76$V_{min}$) than dense obstacle (0.80$V_{min}$), this is because a more challenging environment brings more complex trajectories for UAVs to follow, which is more critical to bit errors. \textsc{BERRY} is adaptive across environments and consistently achieves improved robustness and system efficiency.

% More challenging environ- ment with a higher density of obstacles is also more difficult for the anomaly detection and recovery schemes to recover from errors. For the Dense environment, a MAV has more complex trajectories to follow and more dynamic actions in response to the obstacles, making the range of the variable dis- tribution wider. The wider distribution increases the number of false-negative detection. Thus, there is still a 20.1% gap between autoencoder-based recovery results and golden for the worse case. On the other hand, for the obstacle-free Farm environment or Sparse, the autoencoder-based technique can achieve a similar performance as the golden run.

\subsection{UAV Platform and Policy Architecture Evaluation}
\label{subsec:drone_type}
\begin{figure}
% \vspace{-6pt}
\begin{minipage}[b]{\linewidth}
    \centering
    \includegraphics[width=\columnwidth]{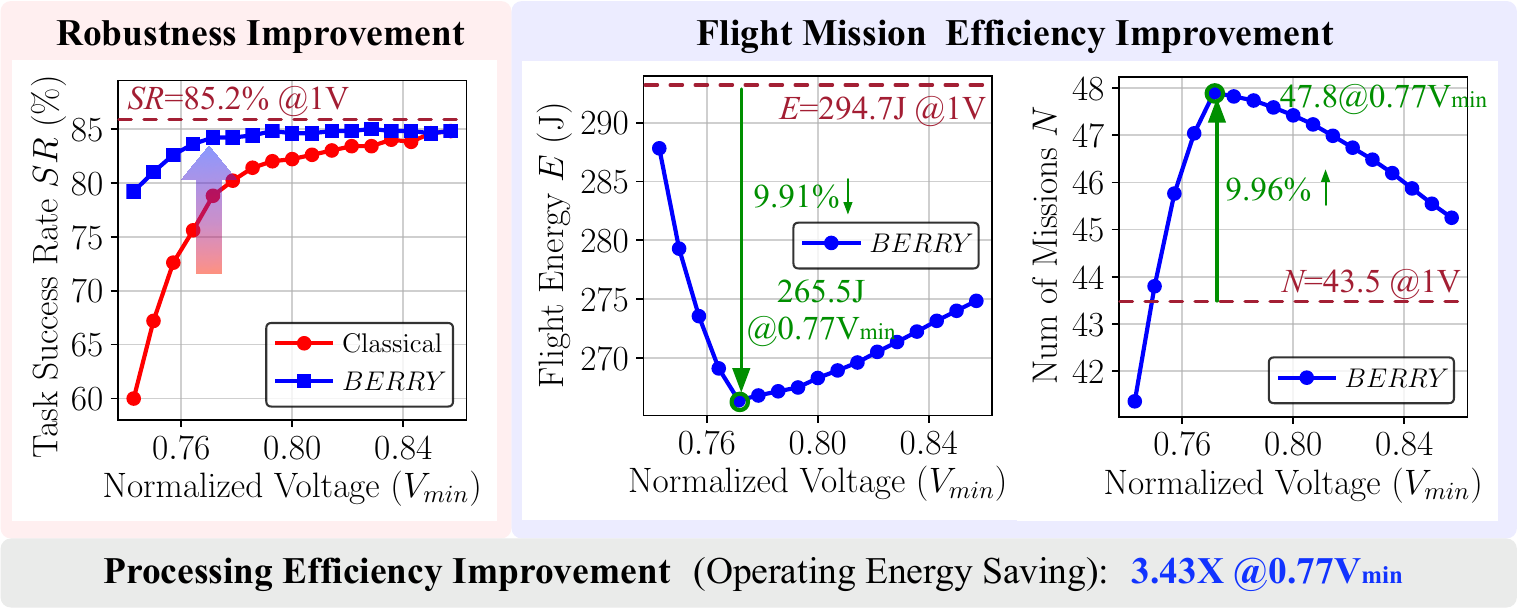}
\end{minipage}% 
 \vspace{0.1in}
\begin{minipage}[b]{\columnwidth}
\small
    \renewcommand*{\arraystretch}{1.25}
\resizebox{\linewidth}{!}{%
\begin{tabular}{c|c|cc|cc}
\hline
\textbf{\begin{tabular}[c]{@{}c@{}}UAV\\ Type\end{tabular}} & \textbf{\begin{tabular}[c]{@{}c@{}}Network\\ Policy\end{tabular}} & \multicolumn{1}{c}{\textbf{\begin{tabular}[c]{@{}c@{}}Rotor\\ Power\end{tabular}}} & \textbf{\begin{tabular}[c]{@{}c@{}}Compute\\ Power\end{tabular}} & \textbf{\begin{tabular}[c]{@{}c@{}}\textsc{BERRY}\\Flight Energy $\downarrow$\end{tabular}} & \textbf{\begin{tabular}[c]{@{}c@{}}\textsc{BERRY} \\ \#Missions $\uparrow$\end{tabular}} \\ \hline
Crazyflie                                                     & C3F2                                                              & 93.5\%                                                                             & 6.5\%                                                            & 15.62\%                                                                 & 18.51\%                                                                 \\ \hline
DJI Tello                                                     & C3F2                                                              & 97.2\%                                                                             & 2.8\%                                                            & 9.91\%                                                                  & 9.96\%                                                                  \\ \hline
DJI Tello                                                     & C5F4                                                              & 94.9\%                                                                             & 4.1\%                                                            & 13.12\%                                                                 & 14.38\%                                                                 \\ \hline
\end{tabular}
}
\captionof{figure}{\textbf{Effectiveness across Different UAVs and Models}. \textsc{BERRY} is evaluated on two UAVs and two models (top: DJI Tello with C3F2), and consistently improves robustness and efficiency. Higher processing power ratio makes \textsc{BERRY} bring more mission improvements.}
\label{fig:diff_drone}
\end{minipage}
\end{figure}

\textbf{Effectiveness across UAV Platforms.} In Fig.~\ref{fig:diff_drone}, we evaluate \textsc{BERRY} on another UAV platform, DJI Tello (Sec.~\ref{subsec:setup}), with the same C3F2 autonomy policy. DJI Tello has a larger frame size and takeoff weight than Crazyflie, thus the rotor power accounts for a higher ratio of total power (97.2\%). Even with a smaller compute power ratio (2.8\%), \textsc{BERRY} still consistently improves success rate under various low voltage levels, and achieves 9.91\% lower flight energy and 9.96\% more missions at 0.77$V_{min}$ with 3.43$\times$ processing efficiency.

\textbf{Effectiveness across Model Architecture.} In Fig.~\ref{fig:diff_drone}, we also evaluate \textsc{BERRY} on another autonomy policy architecture C5F4 (5 Conv and 4 FC layers) on DJI Tello UAV. C5F4 architecture has 1.98$\times$ parameters than C3F2 and accounts for 4.1\% of total power. \textsc{BERRY} saves up to 13.12\% flight energy usage and increases the number of missions by 14.38\%. Higher compute power attributes enable \textsc{BERRY} to bring more system-level benefits, and \textsc{BERRY} consistently improves task robustness and efficiency across various UAVs and models.

\begin{table}[t!]
\scriptsize
\centering
\caption{\textbf{Effectiveness across Different Profiled Bit Errors.} 
\textsc{BERRY} is evaluated on different profiled chips including column-aligned error distributions. \textsc{BERRY} generalizes well across chips and voltages, with robustness and efficiency improvements.}
\renewcommand*{\arraystretch}{1.15}
\setlength\tabcolsep{4pt}
\resizebox{\linewidth}{!}{%
\begin{tabular}{l|cc|cc}
\hline
Chips and Error Rates $p$ (\%)                   & \multicolumn{2}{c|}{SuccRate $SR$ (\%)} & \multicolumn{2}{c}{Flight Energy $E$ (J)} \\ \hline
\textbf{Chip 1} (random  pattern)  & $p$=0.16           & $p$=0.74          & $p$=0.16             & $p$=0.74          \\
\textsc{BERRY} $p$=0.5        & $SR$=84.0               & $SR$=77.2            & $E$=48.46              & $E$=54.63            \\ \hline
\textbf{Chip 2} (column-aligned) & $p$=0.067          & $p$=0.32          & $p$=0.067            & $p$=0.32         \\ 
\textsc{BERRY} $p$=0.5        & $SR$=86.0               & $SR$=81.8            & $E$=46.98              & $E$=51.27            \\ \hline
Baseline $p$=0 @1$V$                  & \multicolumn{2}{c|}{$SR$=88.4} & \multicolumn{2}{c}{$E$=53.19} \\ \hline
\end{tabular}
}
\label{tab:gen_chip}
% \vspace{-3pt}
\end{table}

% \begin{table}[t!]
% \scriptsize
% \centering
% \caption{\textbf{Effectiveness on Different Profile Bit Errors.} 
% \textsc{BERRY} is evaluated on two different profiled chips. Chip 2 exhibits a greatly different bit error distribution where bit errors are strongly aligned along columns and biased towards 0-to-1 flips. \textsc{BERRY} generalizes well on both chips and improves robustness and flight efficiency.}
% \renewcommand*{\arraystretch}{1.05}
% \resizebox{\linewidth}{!}{%
% \begin{tabular}{l|cc|cc}
% \hline
%                    & \multicolumn{2}{c|}{SuccRate $SR$ (\%)} & \multicolumn{2}{c}{Flight Energy $E$ (J)} \\ \hline
% \textbf{Chip 1} (Fig.~\ref{fig:vol_ber_energy} left)  & $p$=0.16\%           & $p$=0.74\%          & $p$=0.16\%             & $p$=0.74\%          \\
% \textsc{BERRY} $p$=0.5\%        & $SR$=84               & $SR$=77.2            & $E$=48.46              & $E$=54.63            \\ \hline
% \textbf{Chip 2} (Fig.~\ref{fig:vol_ber_energy} right) & $p$=0.067\%          & $p$=0.32\%          & $p$=0.067\%            & $p$=0.32\%         \\ 
% \textsc{BERRY} $p$=0.5\%        & $SR$=86               & $SR$=81.8            & $E$=46.98              & $E$=51.27            \\ \hline
% Baseline $p$=0\% @1$V$                  & \multicolumn{2}{c|}{$SR$=88.4} & \multicolumn{2}{c}{$E$=53.19} \\ \hline
% \end{tabular}
% }
% \label{tab:gen_chip}
% \vspace{-3pt}
% \end{table}

\subsection{Bit Error Pattern Evaluation}
\label{subsec:pattern}
\textbf{Effectiveness across Profiled Bit Errors.} In Tab.~\ref{tab:gen_chip}, we evaluate \textsc{BERRY} on two different profiled bit error patterns from test chips in ~\cite{danteHPCA2019,eatMLSYS2021}  (Fig.~\ref{fig:vol_ber_energy}), one showing a random spatial error pattern and another showing a column-aligned pattern with a bias towards 0-to-1 flips. \textsc{BERRY} generalizes well to both lower and higher bit error rates than trained on, showing robustness and mission efficiency improvements.
% , as well as other row-aligned and center-gathering bit error patterns
% (Profiling is done at different voltage levels, resulting in various bit error rates. To simulate various weights to memory mappings, we apply various offsets before linearly mapping weights to the profiled SRAM arrays.) 
% Evaluation \textsc{BERRY} on other profiled fault patterns (row-aligned and center-gathering) shows consistent observation.
% also evaluate on more column aligned, row aligned, 4-5 different fault models~\cite{}, for different fault models BERRY work. don't have to from specifc chip, but can generate fault model, maake it more general.
% We need to show generalization because not all drones can support online training.

 % \vspace{-0.02in}
\subsection{On-Device Error-Aware Robust Learning}
\textbf{Effectiveness of On-Device Robust Learning at Lower Voltage and Better Quality-of-Flight.}
On-device fine-tuning is needed in some scenarios for UAV adapting to environments~\cite{anwar2020autonomous}. \textsc{BERRY} framework supports on-device robust learning where the UAV can learn the bit errors directly at low-voltage chips (Sec.~\ref{sec:training}). While on-device learning consumes on-the-fly energy, compared to offline \textsc{BERRY}, the UAV can enable lower operating voltage and improved robustness due to the same fault pattern in learning and inference. The achieved lower voltage can save more flight energy usage for further tasks. 
As in Tab.~\ref{tab:online_berry}, with 6k on-device training steps, the Tello UAV achieves robust fly under 0.70$V_{min}$, resulting in 13.41\% less flight energy with 4.16$\times$ less operating energy than 1$V$ operation, and 3.89\% less flight energy than offline \textsc{BERRY}. 
Since not all UAVs support on-device training, with the inherent tradeoff between learning-consumed energy and model efficiency, \textsc{BERRY} framework provides the flexibility for offline or on-device robust learning based on scenarios.

\begin{table}[t!]
% \huge
% \vspace{-5pt}
\centering
\caption{\textbf{On-Device Error-Aware Robust Learning.} Learning the bit errors directly on low-voltage device enables lower operating voltage and improved robustness, resulting in more flight energy savings, while with the cost of on-the-fly learning energy consumption.}
\renewcommand*{\arraystretch}{1.15}
\setlength\tabcolsep{1.5pt}
\resizebox{\linewidth}{!}{%
\begin{threeparttable}
\begin{tabular}{lccc|c|c|cc}
\hline
\multicolumn{4}{c|}{\textbf{Low-Voltage Operation}}                                                                                                                                                                                                                                & \textbf{\begin{tabular}[c]{@{}c@{}}Operating\\ Efficiency\end{tabular}} & \textbf{Robustness}                                              & \multicolumn{2}{c}{\textbf{Quality-of-Flight}}                                                                                    \\ \hline
\multicolumn{2}{c|}{\begin{tabular}[c]{@{}c@{}}Num. of \\Learning Steps\end{tabular}}                                                & \begin{tabular}[c]{@{}c@{}}Operating\\ Voltage \end{tabular} & \begin{tabular}[c]{@{}c@{}}Learning\\ Energy (J)\end{tabular} & \begin{tabular}[c]{@{}c@{}}Energy\\ Savings\end{tabular}                & \begin{tabular}[c]{@{}c@{}}Success \\ Rate (\%)\end{tabular} & \begin{tabular}[c]{@{}c@{}}Flight \\ Energy (J)\end{tabular} & \begin{tabular}[c]{@{}c@{}}Num. of\\ Missions\tnote{*}\end{tabular} \\ \hline
\multicolumn{1}{c|}{\multirow{4}{*}{\begin{tabular}[c]{@{}l@{}}On-Device\\ \textsc{BERRY}\end{tabular}}} & \multicolumn{1}{c|}{\multirow{2}{*}{4000}} & 0.77$V_{min}$                                                                & 1849                                                            & 3.43$\times$                                                                   & 84.6                                                           & 264.2                                                            & 48.19                                                         \\
\multicolumn{1}{c|}{}                                                                        & \multicolumn{1}{c|}{}                      & 0.70$V_{min}$                                                                & 1807                                                            & 4.16$\times$                                                                   & 82.4                                                           & 266.5                                                            & 46.52                                                         \\ \cline{2-8} 
\multicolumn{1}{l|}{}                                                                        & \multicolumn{1}{c|}{\multirow{2}{*}{6000}} & 0.77$V_{min}$                                                                & 2775                                                            & 3.43$\times$                                                                   & 85.0                                                           & 260.9                                                            & 49.03                                                         \\
\multicolumn{1}{l|}{}                                                                        & \multicolumn{1}{c|}{}                      & 0.70$V_{min}$                                                                & 2711                                                            & 4.16$\times$                                                                   & 84.8                                                             & \textbf{255.1}                                                            & \textbf{50.01}                                                         \\ \hline
\multicolumn{2}{c|}{\multirow{2}{*}{\begin{tabular}[c]{@{}c@{}}Offline \\ \textsc{BERRY}\end{tabular}}}                                                                                                        & 0.77$V_{min}$                                                                & -                                                                & 3.43$\times$                                                                   & 84.4                                                           & 265.5                                                            & 47.84                                                         \\
\multicolumn{2}{l|}{}                                                                                                                     & 0.70$V_{min}$                                                                & -                                                                & 4.16$\times$                                                                   & 63.8                                                           & 375.6                                                            & 25.56                                                         \\ \hline
\multicolumn{2}{c|}{Baseline}                                                                                                             & 1$V$                                                                   & -                                                                & 1$\times$                                                                      & 85.2                                                           & 294.7                                                            & 43.50                                                         \\ \hline
\end{tabular}
\begin{tablenotes}
        \item[*] \scriptsize Does not include on-device learning flight energy, evaluated for missions after learning.
      \end{tablenotes}
    \end{threeparttable}
}
\label{tab:online_berry}
% \vspace{-1pt}
\end{table}

\section{Conclusion}
\label{sec:conclusion}
% This paper systematically investigated the performance of  UAV systems when lowering operating voltages. A novel robust learning framework, \textsc{BERRY}, was proposed to improve UAVs' robustness, efficiency and quality-of-flight. The effectiveness of \textsc{BERRY} was validated by the significant improvement observed in 72 deployment scenarios.

\textsc{BERRY} is a promising robust learning framework unlocking practical low-voltage operation advantages on RL-enabled autonomous systems. \textsc{BERRY} relies on the systematic discovery of relationship between voltage and mission performance, and supports both offline and on-device error-aware learning. We have demonstrated that \textsc{BERRY} consistently improves task robustness, operating efficiency, and mission performance and achieves up to 15.62\% energy savings, 18.51\% increase in successful missions with 3.43$\times$ processing energy reduction across environments, UAVs, and autonomy policies. We anticipate \textsc{BERRY} framework being useful in exploring robust and efficient low-voltage operations in other autonomous systems.
% \vspace{-12pt}
\section*{Acknowledgements}
We would like to thank Srivatsan Krishnan from Harvard for helpful discussions. This work was supported by CoCoSys, one of the seven centers in JUMP2.0, a Semiconductor Research Corporation (SRC) program sponsored by DARPA; IARPA MicroE4AI program; and CRNCH PhD Fellowship.
\bibliographystyle{ieeetr}
\bibliography{refs}
%%%%%%%%%%%%%%%%%%%%%%%%%%%%%%%%%%%%

\end{document}